# Deep learning for state estimation of commercial sodium-ion batteries using partial charging profiles: validation with a multi-temperature ageing dataset

Jiapeng Liu[1,*], Lunte Li[2], Jing Xiang[2], Laiyong Xie[2], Yuhao Wang[3], Francesco Ciucci[4,5,*]

1 School of Advanced Energy, Sun Yat-Sen University, 518107 Shenzhen, China

2 Hong Kong Applied Science and Technology Research Institute, Photonics Centre, 2 Science Park E Ave, Science Park, Hong Kong

3 Department of Mechanical and Aerospace Engineering, The Hong Kong University of Science and Technology, Clear Water Bay, Hong Kong

4 Chair of Electrode Design for Electrochemical Energy Systems, University of Bayreuth, Weiherstrasse 26, 95448 Bayreuth, Bavaria, Germany

5 Bavarian Center for Battery Technology (BayBatt), University of Bayreuth, Universitätsstraße 30, 95447 Bayreuth, Germany

* Corresponding author: liujp59@mail.sysu.edu.cn, francesco.ciucci@uni-bayreuth.de

**Abstract**

Accurately predicting the state of health for sodium-ion batteries is crucial for managing battery modules, playing a vital role in ensuring operational safety. However, highly accurate models available thus far are rare due to a lack of aging data for sodium-ion batteries. In this study, we experimentally collected 53 single cells at four temperatures (0, 25, 35, and 45 °C), along with two battery modules in the lab. By utilizing the charging profiles, we were able to predict the SOC, capacity, and SOH simultaneously. This was achieved by designing a new framework that integrates the neural ordinary differential equation and 2D convolutional neural networks, using the partial charging profile as input. The charging profile is partitioned into segments, and each segment is fed into the network to output the SOC. For capacity and SOH prediction, we first aggregated the extracted features corresponding to segments from one cycle, after which an embedding block for temperature is concatenated for the final prediction. This novel approach eliminates the issue of multiple outputs for a single target. Our model demonstrated an $R^2$ accuracy of 0.998 for SOC and 0.997 for SOH across single cells at various temperatures. Furthermore, the trained model can be employed to predict single cells at temperatures outside the training set and battery modules with different capacity and current levels. The results presented here highlight the high accuracy of our model and its capability to predict multiple targets simultaneously using a partial charging profile.

# 1 Introduction

Rechargeable batteries, particularly lithium-ion batteries (LIBs) and sodium-ion batteries (SIBs), have been widely adopted due to the rapid development of electric vehicles and large-scale energy storage. Compared to lithium counterparts, SIBs show significant potential for energy storage technologies thanks to their abundant raw materials, low-cost, and environmental friendliness[1]. Although the capacity of rechargeable batteries has significantly improved over the decades, they still experience capacity decay during cycling. Therefore, accurately predicting the batteries' state metrics, such as state of charge (SOC) and state of health (SOH), plays a crucial role in managing the battery system[2]. Moreover, SOH directly reflects a battery's remaining useful life, with 80% typically regarded as the threshold for electric vehicles. An accurate assessment of SOH not only aids in optimizing battery usage strategies and maintenance plans but also helps prevent unexpected accidents caused by battery degradation.

To estimate a battery's state during cycling, various methods exist, ranging from direct techniques[3,4] to electrochemical methods[5,6] to the recently popular data-driven approaches[7,8]. The most straightforward method to estimate the SOC or SOH is ampere-hour counting when the current is available. However, this always requires an initial SOC value, and the accumulated error may ultimately result in significant inaccuracy over a long measurement period. Moreover, these simple methods often struggle to accurately capture the complex physical and chemical changes in batteries[9]. Direct techniques, such as scanning electron microscopy and Raman spectroscopy, are useful for observing the microstructure and analyzing deeper degradation mechanisms. However, these experiments are destructive and only suitable for small-scale prototypes. Electrochemical methods, including differential voltage analysis[10], equivalent circuit models (ECM)[11], and porous-electrode-based degradation models[12,13], can effectively capture the dynamic changes in batteries based on physical laws with tuned parameters derived from experiments. Nevertheless, these approaches generally require specific testing conditions or data processing to yield reliable results, complicating the entire process. Thanks to rapid advancements in computational power and the abundance of data, machine learning-based SOH prediction models have attracted substantial attention from both industrial and academic sectors. By leveraging extensive historical operational data—such as voltage, temperature, and current—along with sophisticated algorithms like support vector machines[14,15], Gaussian processes[16-19], long short-term memory (LSTM) networks[20-22], and recurrent neural networks (RNNs)[23-25], these models generally demonstrate better performance compared to traditional ones.

Despite the extensive methods designed for LIBs, models available for SIBs lag far behind. This may be attributed to the scarcity of openly accessible aging datasets for SIBs, while such datasets for LIBs are easily obtainable; *for example*, the Oxford[26] and NASA[27] datasets are the two most commonly used reference datasets tested using commercial LIBs. Despite the similar working principles between the two types of batteries, it is questionable to transfer models developed for LIBs directly to SIBs, considering that the degradation mechanisms may differ for several reasons: (1) the larger radius of $Na^+$ (1.90 Å) *compared to* $Li^+$ (1.67 Å) may lead to greater volume expansion and slower reactions in SIBs, (2) the common electrode materials (*e.g.*, hard carbon, $NaVPO_4$) of SIBs show different structures compared to those of LIBs (*e.g.*, graphite, $LiFePO_4$), and (3) the operational electrochemical windows and transport kinetics may exhibit different behaviors. These factors might play a crucial role in reusing models for SIBs from their lithium counterparts. Consequently, there is an urgent need to construct aging datasets and establish a suitable model specifically for SIBs. To address these challenges, we collected aging data from 53 commercial sodium batteries, each with a nominal capacity of 10 Ah, in our laboratory at four different temperatures. We further cycled two additional modules to augment the dataset and verify the applicability of the developed model on battery module packs.

With the collected aging data of commercial sodium batteries and module packs, we propose a data-driven framework capable of predicting both the SOC during different charging periods and the SOH at the cycle endpoint. The framework is based on neural ordinary differential equations and convolutional neural networks. We must stress that the developed model only uses easily accessible data, such as charge/discharge curves and temperature as input, making the model a practical and scalable solution for real-world applications. Unlike electrochemical models that require additional tests such as electrochemical impedance spectroscopy, our approach relies solely on data gathered during the batteries' normal operation. Moreover, our approach is highly flexible, as a complete charging process is not necessary for predicting the corresponding SOH. By harnessing the powerful representation capabilities of deep-learning frameworks, our model demonstrates exceptional prediction accuracy for both SOC and SOH of single cells at various temperatures. Additionally, we tested the model's transferability to battery module packs and the results indicated that the model trained with single-cell data points is able to capture the performance of integrated module pack predictions with high accuracy. The simplicity and transferability of the developed framework enhance its applicability across various industries, particularly for monitoring and optimizing the performance of sodium battery systems in commercial settings.

**2 Dataset**

Commercial cylindrical sodium batteries with a specified radius of 33.2 mm and height of 140.3 mm were purchased for testing. To investigate the effect of temperature on battery aging, we divided the batteries into several groups, and each group was tested at a specific temperature controlled by the experiment. During cycling, the corresponding information on current, voltage, and capacity was recorded using the LAND system. All batteries were charged following a similar protocol, where a constant current charging (CC) process is followed by a constant voltage charging (CV). In the CC phase, the battery was charged from ~2.15 V to ~4 V with a current of 5 A, whereas in the CV phase, the battery was further charged at 4 V until the current gradually decreased to 0.5 A. Due to varying testing conditions, batteries at different temperatures may experience different voltage ranges in the CC phase, as detailed in Table 1. The time-dependent evolution of voltage and current profiles during the charging process is illustrated in Figure 1. As one can observe, with an increasing cycle number, the time required to charge to 4 V in the CC phase decreases, indicating capacity decay after cycling the battery.

Table 1 Available maximum cycle numbers and specific charging information during the CC and CV for all cells tested at different temperatures.

| Temperature | Battery label | Max cycle number[§] | V range of CC | I range of CV |
|---|---|---|---|---|
| 0 °C | 0-B2-1 to 0-B2-4 | ~ 800 | ~2.3 to 3.9 V | 5 to 0.5 A |
| | 0-B2-5 to 0-B2-12 | ~ 250 | | |
| 25 °C | 25-B1-1 to 25-B1-8 | ~1040 | ~2.18 to 4.0 V | |
| | 25-B2-1 to 25-B2-8 | ~580 | | |
| | 25-B2-9 to 25-B2-11 | ~350 | | |
| 35 °C | 35-B2-1 to 35-B2-3 | ~390 | ~2.15 to 4.0 V | |
| 45 °C | 45-B1-1 to 45-B1-4 | ~700 | ~2.12 to 4.0 V | |
| | 45-B1-5 to 45-B1-7 | ~390 | | |
| | 45-B1-8 | 66 | | |
| | 45-B2-1 to 45-B2-8 | ~370 | | |
| | 45-B2-9 to 45-B2-11 | ~240 | | |

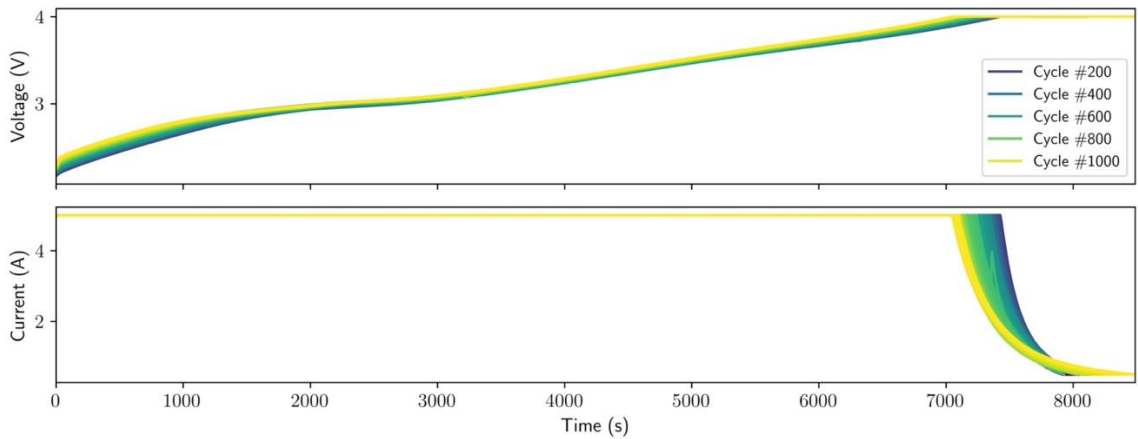

Figure 1 Time-dependent voltage and current profiles during the charging process at different cycles for the battery 25-B1-8 tested at 25 °C.

---

[§] The maximum cycle number at each temperature may vary based on the starting time, with an approximation provided for cells with nearly identical available cycles.

The recorded capacity *versus* cycle number at different temperatures is illustrated in Figure 2. As the capacity slightly increases during the initial cycling, we will define the SOH of the k[th] using the 3[rd] cycle as a reference, namely,

$$SOH_k = \frac{Q_k}{Q_3} \tag{1}$$

where $Q_k$ and $Q_3$ represent the discharge capacity of $k$-th cycle and the 3[rd] cycle, respectively. We use the 3[rd] cycle for normalization since the first two cycles are considered activation cycles. More detailed information on the test conditions for all 53 single cells is provided in Table 1. As the batteries were tested in subsequent two batches, the available cycles vary for different batches as well as temperature. The total available cycles for all single cells are also listed in Table 1.

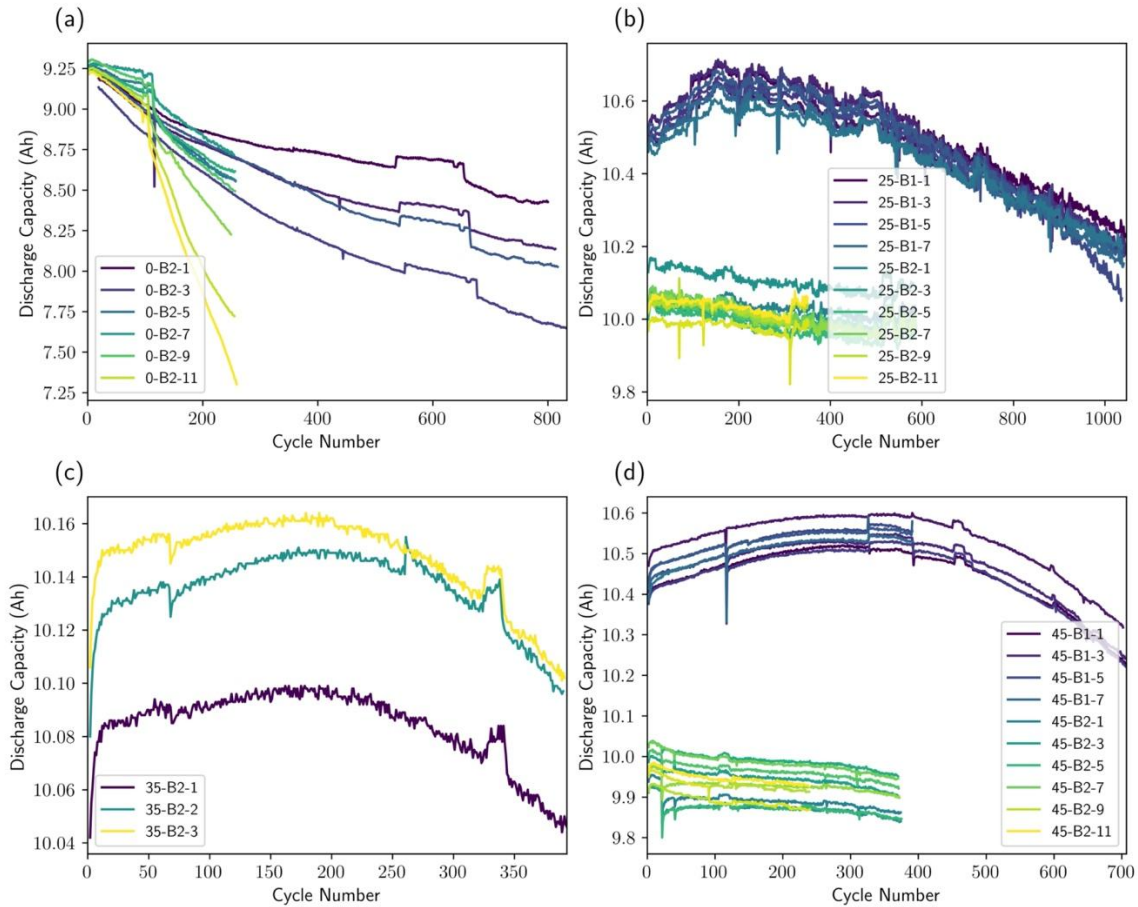

Figure 2 Capacity decay profiles of all 53 single batteries tested at different temperatures, (a) at 0 °C, (b) at 25 °C, (c) at 35 °C, and (d) at 45 °C. Selected battery labels are shown inset the figure

for simplicity. Outliers, in specific, cycles 105-106 for battery 45-B2-6 and cycle 91 for battery 45-B2-9, were removed to better fit the profile.

## 3 Methods

### 3.1 Overview

Different from the ageing test conducted in the laboratory, the load of batteries may vary significantly in practical usage, making the discharging profile out of control. On the other hand, the charging protocol is generally independent of the usage, thus providing consistent profiles across different cycles. Nonetheless, one may encounter the scenario that the battery is partially charged starting with an initial SOC larger than 0% and ending with a SOC smaller than 100%. For such a case, the feature engineering derived using the full charging curve may fail. To address these shortcomings, we propose to use the partial charging data as the input feature to predict both the coming SOC, SOH, and capacity corresponding to the "full cycle" even if it may not be experienced.

Given the collected current and voltage data during cycling, we concatenate them to form a 2-dimensional vector, thus making full use of the charging information. Considering that the current and voltage vary in different ranges, see Figure 1 (b), we first re-scale the current and voltage into the range of [0, 1] following the equations

$$\widetilde{V} = \frac{V - 2.15}{4 - 2.15} \tag{2a}$$

$$\widetilde{I} = \frac{I - 0.5}{5 - 0.5} \tag{2b}$$

where the scaled $\widetilde{V}$ and $\widetilde{I}$ are used as input. In accordance with the partial charging data, we propose a sliding window strategy to split the full charging curve into continuous segments, each with `window_size-1` data points. To make sure the sampling frequencies of all charging curves are consistent, we first fit the data using a time series with a period of 30 seconds, which is also the average sampling frequency of the raw data. Hence, each segment spans `(window_size-1)×30` seconds. In this work, we choose the `window_size` as 128. Moreover, the charging curves are prepended with constant values (2.15 V for the raw voltage and 0.5 A for the raw current) before scaling using (2). This is equivalent to prepending zeros for the scaled current and voltage. An illustration of the sliding window used to construct the input features is shown in Figure 3. For example, the charge profile of $k$-th cycle with $N_k$ sampling points can be split into $N_k$ segments as input features, and the $s$-th input feature $\widetilde{F}_{k,s}$ is composed like

$$\widetilde{V}_{k,s} = \left[\widetilde{V}_{k,s-window\_size+1}, \cdots, \widetilde{V}_{k,s-1}, \widetilde{V}_{k,s}\right] \tag{3a}$$

$$\widetilde{I}_{k,s} = \left[\widetilde{I}_{k,s-window\_size+1}, \cdots, \widetilde{I}_{k,s-1}, \widetilde{I}_{k,s}\right] \tag{3b}$$

$$\widetilde{F}_{k,s} = concat(\widetilde{V}_{k,s}; \widetilde{I}_{k,s}) \tag{3c}$$

where $\widetilde{F}_{k,s} \in \mathbb{R}^{2 \times window\_size}$ and $1 \leq s \leq N_k$. Consistently, the corresponding target with respect to $\widetilde{F}_{k,s}$ is constructed as $(SOC_{k,s}; SOH_k; Q_k)^\top$. By prepending zeros to both $\widetilde{V}$ and $\widetilde{I}$, the targeted SOCs thus fall in a range between 0% and 100%.

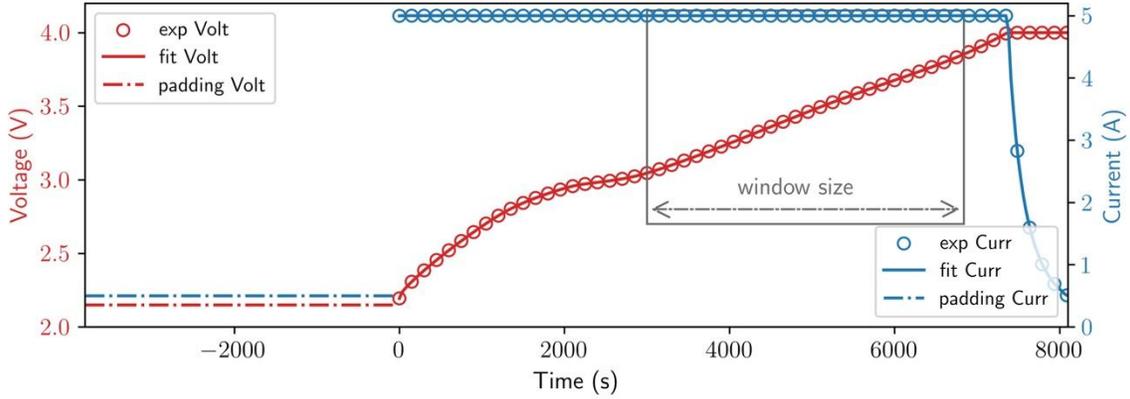

Figure 3 Illustration of sliding window to construct the input feature in this work. The raw current and voltages are first fitted using a fixed sampling frequency of 30s and then prepended with constant values of `window_size-1` length to incorporate the early charging stage.

**3.2 Model**

Given the constructed input feature $\widetilde{F}_{k,s}$, we propose to use the convolutional neural network (CNN) to extract deeper features. Moreover, we will combine the 2-dimensional CNN with the neural ordinary differential equation (NODE)[28,29], which has been previously implemented on the ageing dataset of LIBs[29] and demonstrated the capability of predicting the SOH with higher accuracy. Inspired by this, we will append the NODE block after the CNN block to refine the extracted features. The overall architecture of the model is depicted in Figure 4. To fit the CNN architecture, the feature $\widetilde{F}_{k,s}$ is transformed to $X_{k,s}^{in} \in \mathbb{R}^{1 \times 2 \times window\_size}$ by adding an extra dimension. The model includes a `Conv2D` block, which converts $X_{k,s}^{in}$ into $X_{k,s}^{conv} \in \mathbb{R}^{64 \times 2 \times 4}$. For simplicity, the batch size dimension is omitted. The `NODE` block further refines features while preserving their dimensions. In the `Avg-Pool` layer, the feature is compressed into $X_{k,s}^{pool} \in \mathbb{R}^{64 \times 2 \times 1}$ and then reshaped into $X_{k,s}^{pool} \in \mathbb{R}^{128}$. To predict the SOC, a fully connected layer with 64 hidden units and an output of 1 is applied to $X_{k,s}^{pool} \in \mathbb{R}^{128}$.

For predicting SOH or capacity, we may follow a similar approach to SOC prediction. However, this may lead to reduced performance since all $N_k$ segments from the $k$-th cycle should share the same SOH or capacity target. To address this, we adopt a strategy from crystal graph neural networks[30], where atomic features in a crystal are aggregated to predict material properties. Similarly, we aggregate features from segments of the same cycle into a single representation, $X_k^{agg} \in \mathbb{R}^{128}$, by averaging:

$$X_k^{agg} = \underset{s \in k}{\text{mean}}\, X_{k,s}^{pool} \tag{4}$$

where $s \in k$ selects $X_{k,s}^{pool}$, the feature of the $s$-th segment from the $k$-th cycle. This operation is akin to average pooling and can be implemented using PyTorch's `pooling` function. Notably, equation (4) allows flexibility in omitting some segment features, enabling the model to handle cases where only partial charging curves are available. Additionally, dropping some $X_{k,s}^{pool}$ can enhance the generalizability of $X_k^{agg}$, even when the full charging profile is provided.

After aggregation, we append the temperature feature $T_{num}$ to $X_k^{agg}$ to account for temperature effects on SOH and capacity. $T_{num}$ is encoded using equal-distance discretization (EDD) as follows:

$$T = \frac{T - T_{min}}{T_{max} - T_{min}} \tag{5a}$$

$$\text{EDD}(T) = \text{Embedding}(\lfloor T \times N_T \rfloor) \tag{5b}$$

$$T_{num} = concat(\text{EDD}(T); \text{FFN}(T)) \tag{5c}$$

where Embedding(·) is the Pytorch `embedding` function, $N_T$ is the number of bins, and $T_{max}$ and $T_{min}$ are the temperature range limits (45 °C and 0 °C, respectively). FFN(·) is a feedforward network that applies a nonlinear transformation to $T$. Finally, the concatenated feature is fed into a fully connected layer with 64 hidden units and an output of 2, predicting $SOH_k$ and $Q_k$.

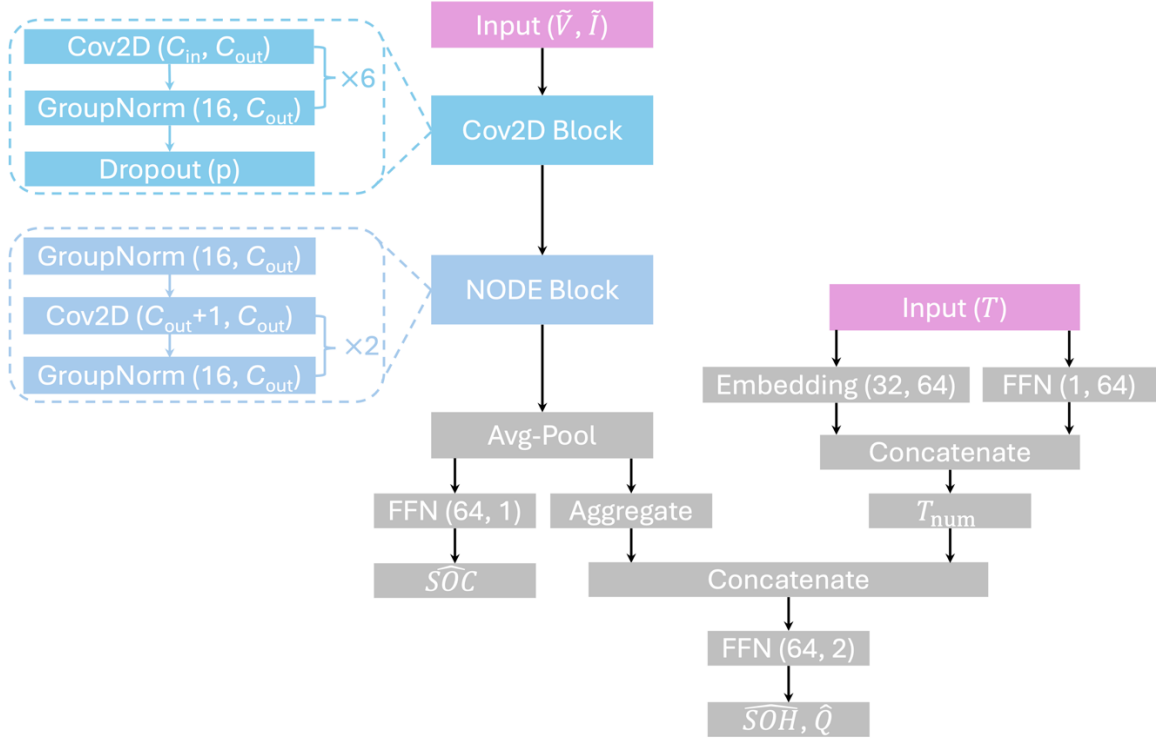

Figure 4 Model architecture proposed in this project to predict the SOC as well as SOH and capacity simultaneously.

To optimize the model, we define the loss function as

$$\mathcal{L} = \frac{\gamma}{N_s} \sum_k \sum_{s=1}^{N_k} \left(SOC_{k,s} - \widehat{SOC}_{k,s}\right)^2 \\ + \frac{1-\gamma}{2M} \sum_k \left[\left(SOH_k - \widehat{SOH}_k\right)^2 + \left(Q_k - \hat{Q}_k\right)^2\right] \tag{6}$$

where the first term represents the mean squared error (MSE) of SOC across all segments of all training cycles, and the second term represents the MSE of SOH and capacity across all cycles. Here, $\widehat{SOC}_{k,s}$ is the predicted SOC for the $s$-th segment of the $k$-th cycle, while $\widehat{SOH}_k$ and $\hat{Q}_k$ are the predicted SOH and capacity of the $k$-th cycle, respectively. The total number of segments in the training dataset is given by $N_s = \sum_k N_k$, and $M$ represents the total number of training cycles.

The parameter $\gamma \in [0,1]$ is a hyperparameter that controls the trade-off between SOC prediction and SOH/capacity prediction. For model evaluation on the validation set, we track the coefficients of determination ($R^2$) for both SOH and capacity. The average metric is calculated as

$$R^2_{avg} = \frac{R^2(Q) + R^2(SOH)}{2} \tag{7}$$

where $R^2(Q)$ and $R^2(SOH)$ are the $R^2$ scores for capacity and SOH, respectively. The model achieving the highest $R^2_{avg}$ is selected as the best to prevent overfitting. Notably, $R^2_{avg}$ does not account for SOC, as our findings indicate that SOC can be accurately predicted using this model.

**4 Results and discussion**

As shown in Table 1, 53 single cells were tested at four different temperatures. To evaluate the model's performance on an unseen temperature, we reserve one temperature (35 °C, 3 cells) for testing, while the remaining three temperatures (0 °C, 25 °C, and 45 °C, 50 cells) are used for training. Since the goal is to predict future SOH or capacity based on historical data, the cycles are split into training, validation, and test subsets in a 70:10:20 ratio. Specifically, for each single cell, the first 70% of cycles are used for training, the next 10% for validation, and the final 20% for testing. Additionally, all cycles tested at 35°C are used to evaluate the trained model. In total, the 50 cells at the three temperatures contain 18,891, 2,695, and 5,431 cycles for training, validation, and testing, respectively. Following the input feature construction method outlined earlier, each charging curve is divided into multiple segments. As a result, the total number of segments for model training, validation, and testing are 4,832,187, 688,115, and 1,392,226, respectively.

We start by reviewing the distribution of the measured target values. The distribution of SOCs for all tested cells at three temperatures (0 °C, 25 °C, and 45 °C) is shown in Figure 5 (a), with the overall mean and standard deviation calculated to be 0.53 and 0.303, respectively. The SOC distribution is nearly uniform, except for a sharp peak around 100%. This peak is primarily caused by the CV phase, where the capacity increases slowly, leading to minimal variation in SOC near 100%. Figure 5 (b) and (c) show the distributions of capacity and SOH for all 50 single cells. A clear multi-modal distribution is observed for capacity, as shown in Figure 5 (b). The profile centered on lower capacities (8-9 Ah) is mainly attributed to cells tested at 0 °C, where lower temperatures reduce capacity. In contrast, the profile centered on higher capacities can be attributed to cells tested in different batches, as shown in Figure 2 (b) and (d). The distribution of SOH, as illustrated in Figure 5 (c), shows less variation between cells tested at different temperatures. However, the SOH distribution has a significant peak near 100%, particularly within the training dataset. This is likely due to the limited number of cycles collected so far, with many batteries not yet experiencing significant degradation. This discrepancy may cause an imbalance between the model's performance on the training and test datasets. The mean and standard deviation of capacity and SOH are [9.95 Ah, 0.711 Ah] and [0.987, 0.032], respectively.

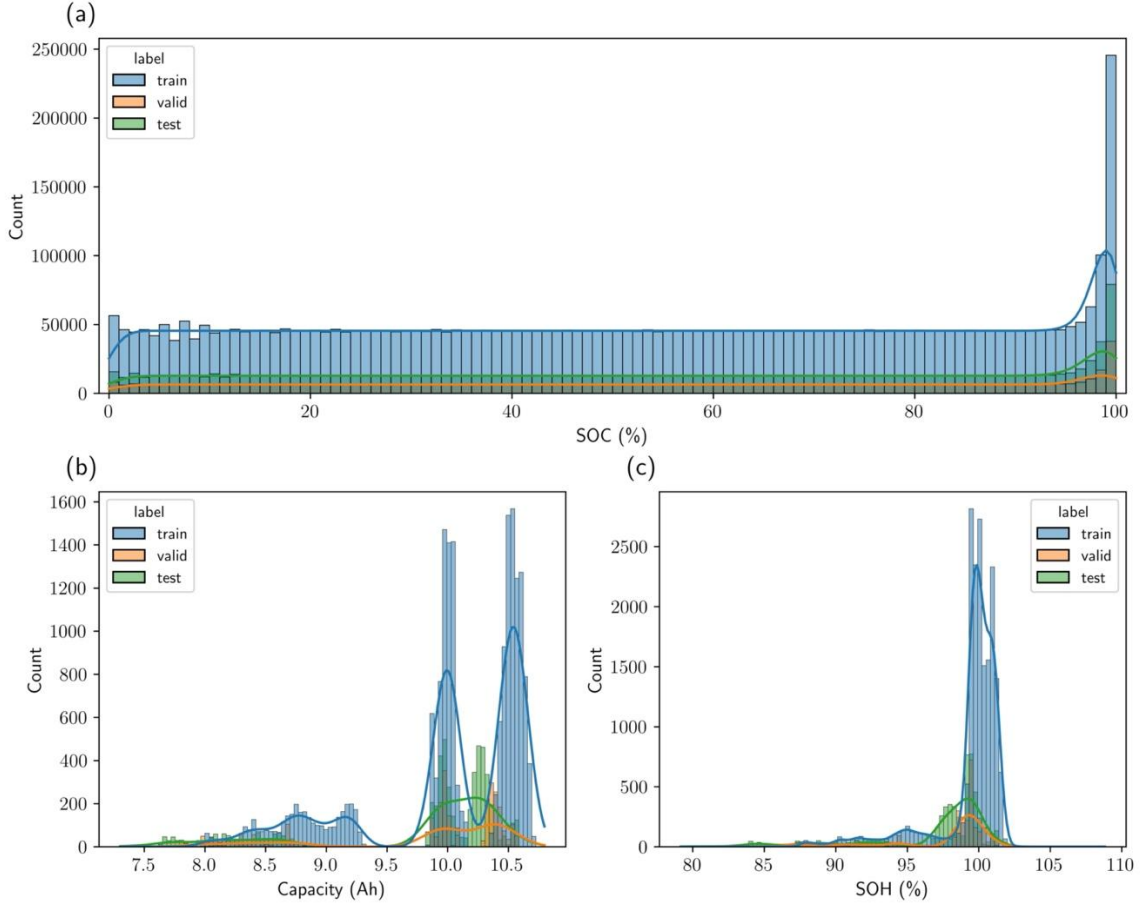

Figure 5 Distribution of (a) SOC corresponding to all segments, (b) capacity, and (c) SOH corresponding to all cycles. The train, valid, test datasets are split by cycle numbers according to the description in the main text.

The model is trained with a batch size of 16, corresponding to 16 cycles. The learning rate starts at $3\times10^{-4}$ and is gradually decreased using the `StepLR` scheduler implemented in PyTorch. The `Adam` optimizer, with a weight decay of $10^{-6}$, is used to optimize the network parameters. During training, we observe that the hyperparameter $\gamma$ in equation (6) has minimal impact on SOC accuracy as long as it is greater than 0. Therefore, we set $\gamma$ to 0.2 to give more weight to minimizing the loss for SOH and capacity. The model is trained for 50 epochs, with the highest $R^2_{\text{avg}}$ consistently achieved around the 30th epoch. The model that achieves the best $R^2_{\text{avg}}$ is then used to predict SOC, SOH, and capacity on the test set.

We first evaluate the model's performance at three temperatures (0, 25, and 45 °C). Figure 6 compares the predicted and measured values for both SOC and capacity. As shown, the predicted SOC values closely align with the measured ones across all temperatures, achieving an $R^2$ of 0.998. Additionally, we present the predicted capacity for the last 20% of cycles from all 50 single cells and compare them with the measured values in Figure 6 (b). It is evident that nearly all data points fall on the $y = x$ line, implying high prediction accuracy. This is further verified by the $R^2$ value of 0.997. Notably, the MSE and mean absolute error (MAE) for capacity are calculated to be 1.92×10$^{-3}$ Ah² and 3.35×10$^{-2}$ Ah, respectively, suggesting that the model effectively captures the relationship between the partial charging curve and cycle capacity. In contrast to previous study[32], where predicted capacities exhibited significant variance, the predictions here demonstrate much better performance. This improvement is primarily attributed to the aggregation of all segment features, as implemented in equation (4).

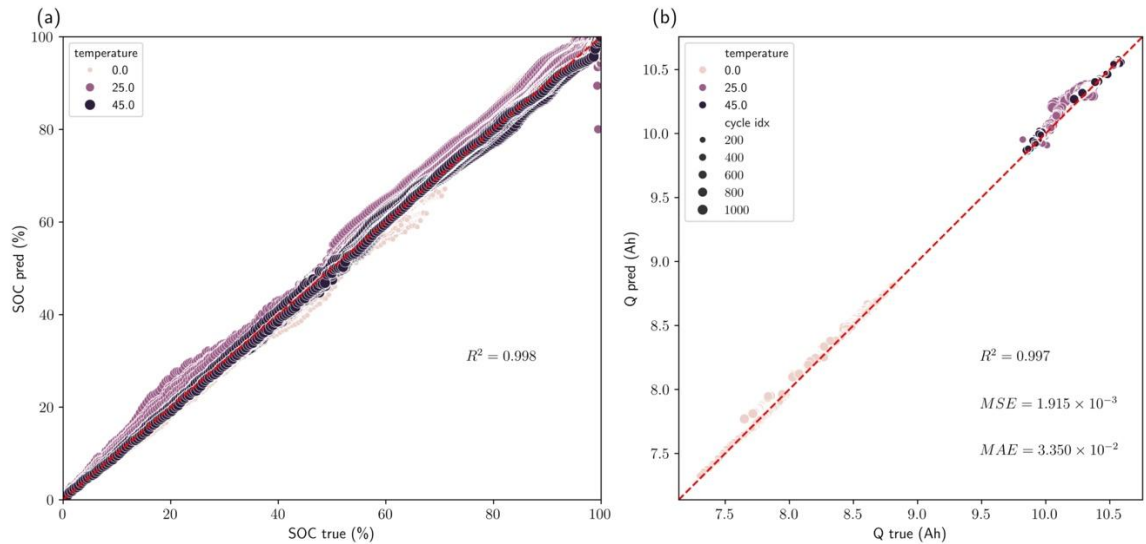

Figure 6 Comparison of prediction *vs* measurement for cells tested at 0 °C, 25 °C, and 45 °C, (a) SOC, (b) capacity.

With the predicted capacity, we can easily calculate the SOH using (1) with respect to the true $Q_3$. We label these calculated SOH values as "cali" to differentiate them from the model's direct output, which will be referred to as "pred." The comparison between $SOH_{\text{cali}}$ and the true values is shown in Figure 7 (a). Due to the variation in $Q_3$ across cells, the calculated $SOH_{\text{cali}}$ performs worse than the predicted capacity, as seen in Figure 6 (b).

However, if we fix $Q_3$ as a constant value for all 50 cells, the $SOH_{\text{cali}}$ should perform equivalently to the predicted capacity. We further compare the results of $SOH_{\text{pred}}$, the model's direct output, with the true values in Figure 7 (b). Surprisingly, the $SOH_{\text{pred}}$ behaves a bit worse than the simply calculated $SOH_{\text{cali}}$, which may be due to the narrower distribution of SOH and the concentration of data points in the training subset, see Figure 5 (b). The model's inferior performance in directly predicting SOH is also reflected in the metrics, where $SOH_{\text{cali}}$ shows a higher $R^2$ value and lower MSE and MAE compared to $SOH_{\text{pred}}$. Nonetheless, both methods exhibit acceptable error ranges, with the maximum error being less than 3%.

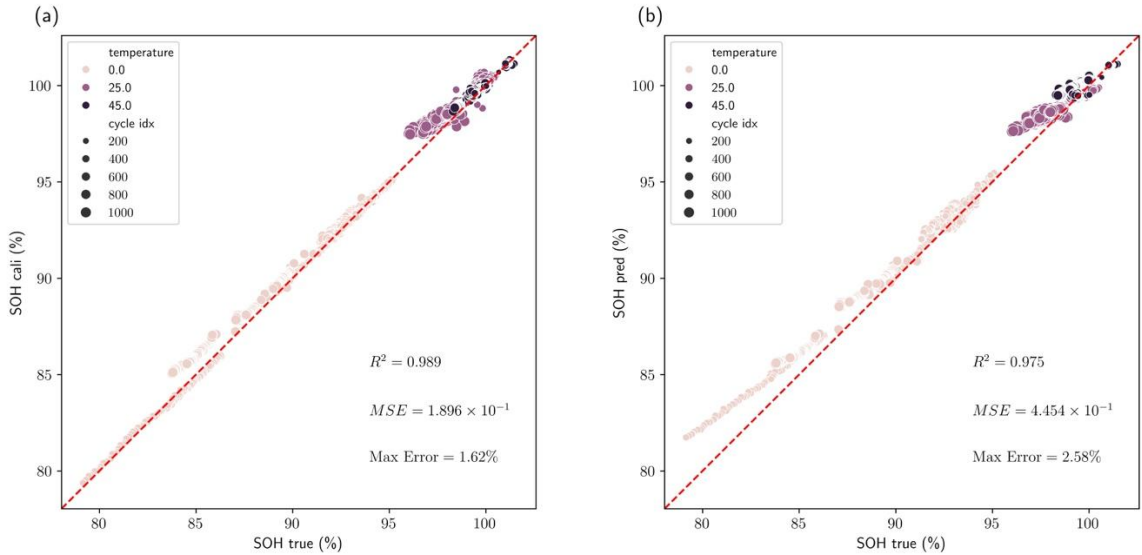

Figure 7 (a) SOH calculated by predicted capacity as shown in Figure 6 *vs* true SOH values. (b) SOH values that are directly predicted by the neural network model *vs* true SOH values. All data points are from cells tested at 0, 25, and 45 °C.

We further evaluate the model's performance by visualizing the predicted SOH alongside the experimentally collected data as a function of the cycle number in Figure 8. The capacity decay profiles differ for cells tested at different temperatures. At lower temperatures, SOH declines significantly over a short cycle life compared to cells tested at higher temperatures. As shown in Figure 8 (a) and (b), the SOH at cycle 800 ($SOH_{800}$) for cells at 0 °C is considerably lower than that for cells at 25 °C. These temperature-dependent capacity decay profiles highlight the challenge of predicting the true state of health.

Using both approaches, we can predict the SOH at cycles beyond those used for training. The model performs perfectly on the training data across all temperatures, indicating that the temperature embedding effectively captures the relationship between SOH and temperature. However, at certain temperatures, the capacity decay pattern may change at different stages of cycling. For instance, for cell 0-B2-1, a notable increase in SOH is observed around cycle 530, followed by a rapid drop after around cycle 640. The differing decay patterns between the training and test sets may complicate model predictions. As illustrated in Figure 8 (a), the $SOH_{\text{cali}}$ aligns more closely with the true SOH compared to the model's direct output, consistent with the results presented in Figure 7. For cells at 25 and 45 °C, the capacity decay profiles differ further, with an initial increase in SOH during the first 200 cycles, as seen in Figure 8 (b) and (d). In line with these observations, the predicted $SOH_{\text{cali}}$ demonstrates better performance than the model's direct output.

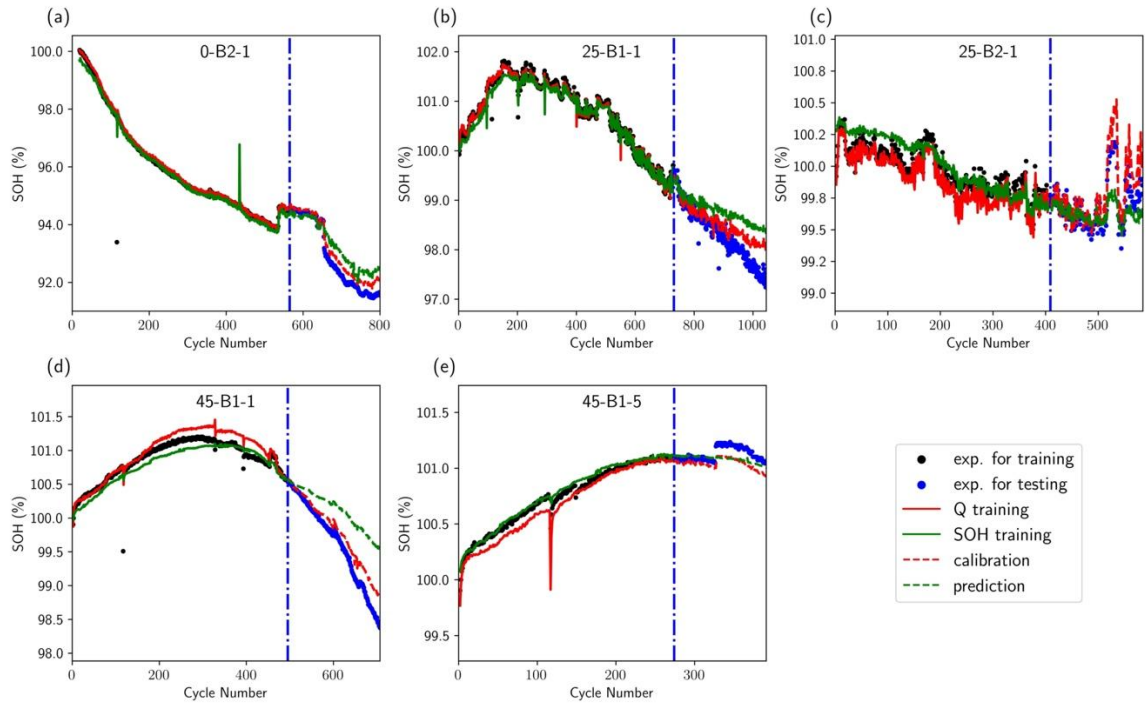

Figure 8 Calculated SOH using either the predicted capacity (labeled as "calibration") or directly from the model (labeled as "prediction") for different cells at (a) 0 °C, (b)-(c) 25 °C, and (d)-(e) 45 °C. The validation set is included in the test set without specifically partitioning in the plot.

The results clearly demonstrate the model's capability in predicting the SOC, SOH, and capacity on the test datasets at three temperatures. We are particularly interested in evaluating the model's transferability to another temperature, 35 °C, which the model has not previously encountered. The predicted SOCs for three cells (35-B2-1 to 35-B2-3) are shown in Figure 9 (a). Despite a small deviation, the predicted SOC aligns well with the true values, achieving a notably high $R^2$ of 0.999, even though the charging profile at 35 °C was never seen during training. This indicates that the model generalizes effectively to other temperatures.

We further calculated the SOH using the predicted capacity, illustrating the results for one cell in Figure 9 (b). For cell 35-B2-1, 390 cycles were tested using the trained model. The calculated SOH matches the true SOH more closely than the model's direct output. Notably, the maximum difference between the prediction and true SOH is less than 1%, which represents a very small margin for SOH prediction. This suggests that the trained model can achieve high accuracy in predicting both capacity and SOH at different temperatures. For the three cells tested at 35 °C, the MSE and MAE for the predicted capacity are calculated as $7.80 \times 10^{-4}$ Ah² and $2.06 \times 10^{-2}$ Ah, respectively, which are even smaller than the metrics for cells tested at 0 °C, 25 °C, and 45 °C, as shown in Figure 6 (b). Moreover, the maximum error for the SOH calculated is observed to be just 0.67% for all three cells, indicating a high level of prediction accuracy. These results suggest that the model performs well across different temperatures.

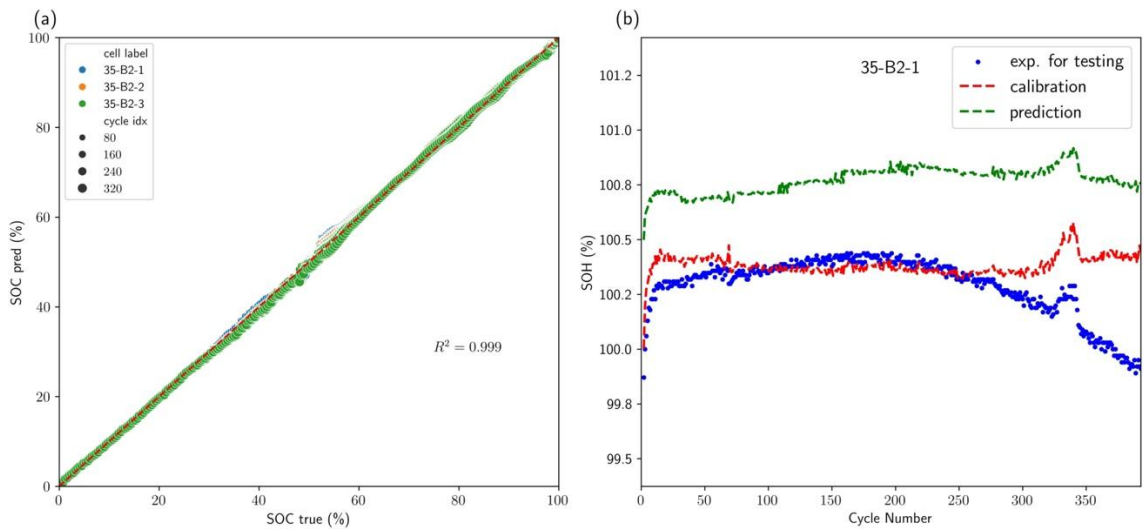

Figure 9 Predicted SOC *vs* measured SOC values for cells tested at 35 °C. Predicted discharged capacity *vs* measured capacity values. (b) SOH by predicted capacity *vs* true SOH values. All data points correspond to cells tested at 35 °C.

The results above demonstrate the trained model's capability to simultaneously predict SOC, SOH, and capacity across various temperatures for a single cell. However, the model's performance on battery module packs has yet to be investigated. In practice, cells are often connected in series (or parallel) to increase output voltage (or current), making the voltage (or current) of a single cell inaccessible. This raises the question of how the trained model will perform in such a setup. To address this, we tested two units, each consisting of four single cells connected in parallel. For each unit, we recorded the total current and voltage. The capacity decay profiles of the two units are illustrated in Figure 10 (a). A noticeable discrepancy appears after approximately 400 cycles in 4p-unit-sample1, primarily due to differences in the testing environments: the first 400 cycles were conducted using the Arbin system, while the subsequent cycles were tested with the LAND system. These environmental differences are also reflected in the charging profiles of each cycle, as shown in Figure 10 (b). Since the training data collected at the three temperatures were obtained using the LAND system, we will exclude the first 400 cycles of 4p-unit-sample1 from the analysis.

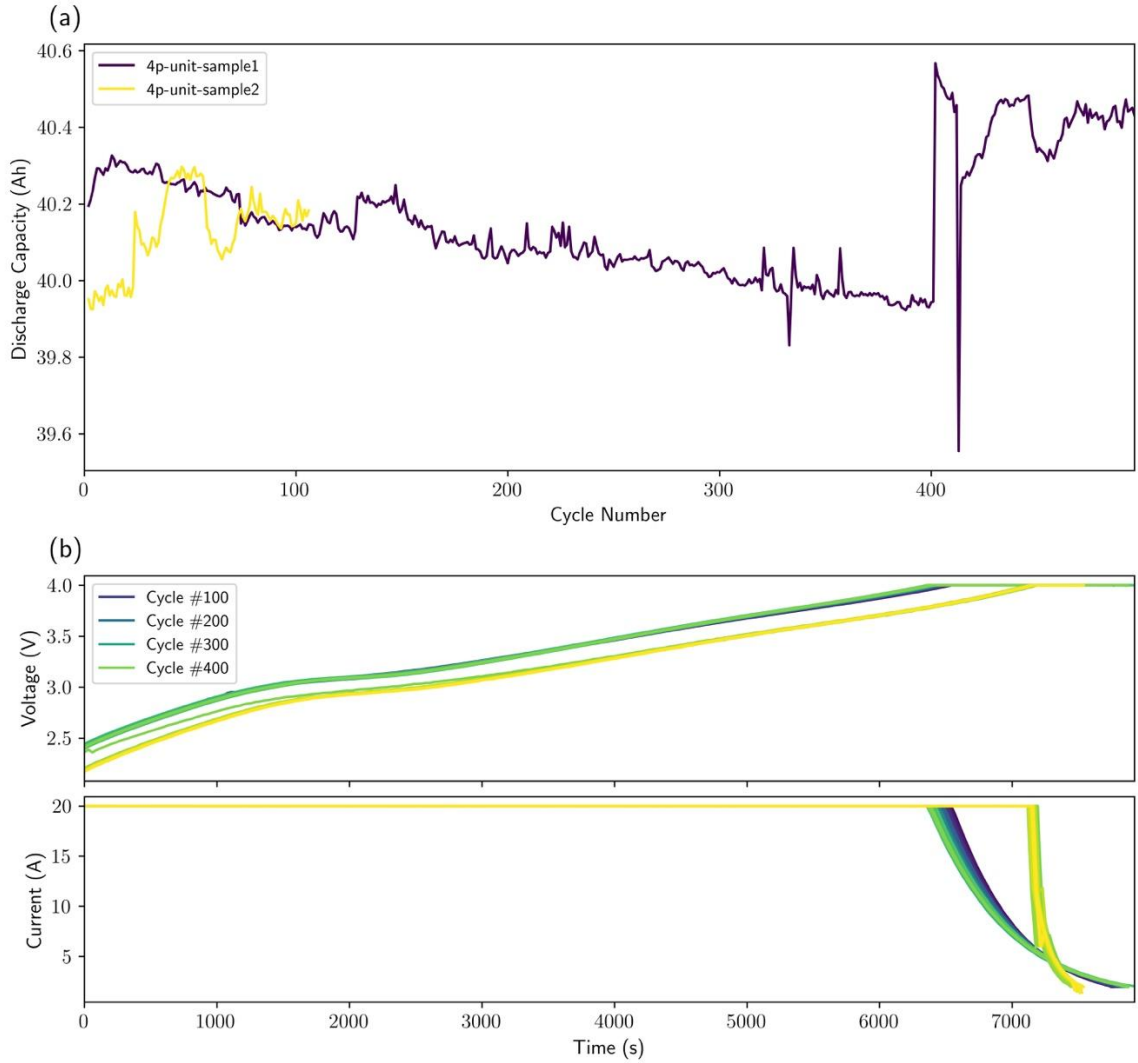

Figure 10 (a) Capacity decay profiles of two modules tested at 25 °C, and (b) Voltage and current profiles of 4p-unit-sample1 at different cycles.

To use the trained model without additional modification, the four cells in the unit are treated as indistinguishable and contribute equally to the unit. Therefore, the capacity and current of the single cell inside are simply one-quarter of the total capacity and current measured, respectively. With this assumption, we can treat each cell in the unit as we did for the cells at 35 °C. The final predicted capacity is four times the model's output. The predicted results of SOC for two units are illustrated in Figure 11 (a). Remarkably, the predicted SOC perfectly matches the true values, achieving an $R^2$ of 1. The predicted capacity values for the two units exhibit high accuracy, with an MSE of $1.38 \times 10^{-2}$ Ah$^2$ and an MAE of $1.14\times10^{-1}$ Ah. Using the accurately predicted capacity, we calculate the $SOH_{\text{cali}}$, which shows a maximum error of just 0.51% compared to the true SOH of the two units. In contrast, the maximum error of the model's direct output, $SOH_{\text{pred}}$, is calculated to be 1.54 %. The predicted SOH for the first unit of 4p-unit-sample1, compared to the true values, is further illustrated in Figure 11 (b), where a closer match between the $SOH_{\text{cali}}$ and experimental data is clearly observed.

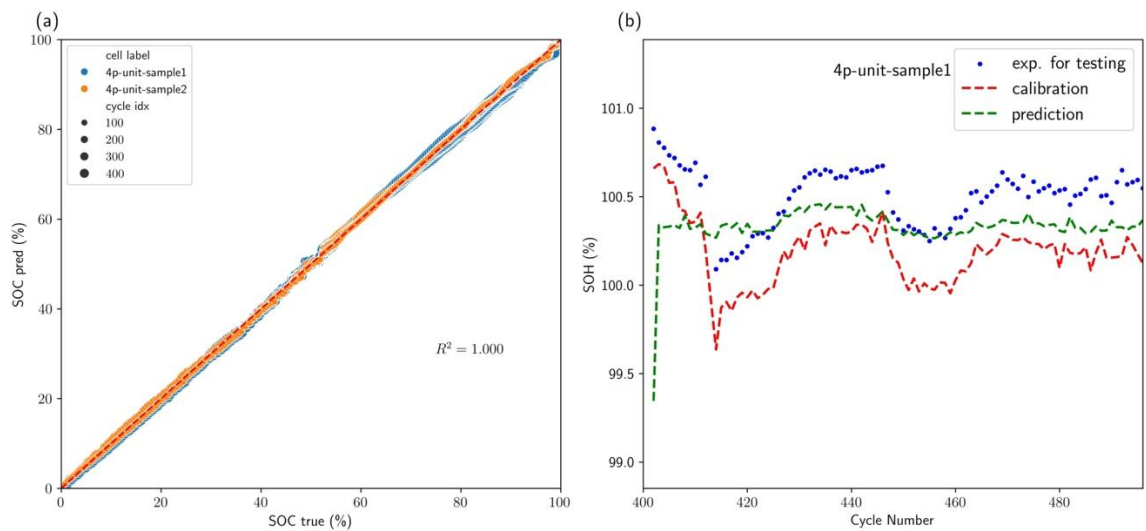

Figure 11 (a) Predicted SOC *vs* true SOC for two modules, (b) predicted SOH *vs* measured values for the first unit of 4p-unit-sample1.

## 5 Conclusion

In this work, we developed a framework based on convolutional neural networks and neural ordinary differential equations to predict the SOC, SOH, and capacity of sodium-ion batteries based on charging profiles. The charging profile is partitioned into different segments of equal length to enhance the training data. These segments are fed directly into the network to learn SOC. For predicting discharge capacity and SOH, features corresponding to segments from the same cycle are aggregated before being passed into the final feedforward network, which outputs capacity and SOH. This approach allows the model to predict both SOC at any point in the charging cycle and the SOH or capacity of the full cycle simultaneously. Additionally, an embedding block is implemented to account for the effect of temperature on the final prediction. The model's performance on SOC, SOH, and capacity is regulated by combining the MSE of all three targets, with a hyperparameter $\gamma$ used to fine-tune the balance. The model was trained on 53 commercial sodium-ion batteries across three temperatures: 0, 25, and 45 °C. Results indicate that the hyperparameter $\gamma$ has a minimal effect on SOC accuracy when $\gamma > 0$, suggesting that SOC is easier to predict within this framework. By monitoring the average regression coefficients of SOH and capacity, the model avoids overfitting the training data.

For all three temperatures, the model demonstrates high accuracy in predicting the SOC of all segments, achieving an $R^2$ value greater than 0.99. This shows the model's strong ability to capture SOC across varying temperatures. Through the designed aggregation method, we achieved an $R^2$ value of 0.997 for capacity predictions on the test dataset across all temperatures. Using the predicted capacity, we were able to recover the $SOH_{\text{cali}}$, which showed a maximum error of 1.62% in the test dataset, demonstrating the model's ability to capture temperature-dependent capacity patterns. With the developed framework, the model can also directly output SOH prediction ($SOH_{\text{pred}}$), though with slightly inferior accuracy compared to $SOH_{\text{cali}}$, with a maximum error of 2.58%. Nevertheless, both predictions fall within an acceptable error range of 3%, which is typically suitable for application. For cells at 35 °C, a temperature the model had not seen during training, we achieved an $R^2$ of 0.999 for SOC and an MAE of 2.06×10$^{-2}$ Ah for capacity. Moreover, the maximum error for $SOH_{\text{cali}}$ was determined to be only 0.67%, even smaller than the error at the trained temperatures, further validating the model's robustness in predicting SOC and SOH across unseen temperatures. Finally, when applied to a battery module, the model achieved an even smaller maximum error of 0.51% for $SOH_{\text{cali}}$.

As highlighted in the model's architecture, it directly predicts SOC, capacity, and SOH. Experimental results confirm that the model can accurately predict SOC from any segment of the charging profile, with high accuracy even for small values of hyperparameter $\gamma$, indicating minimal weight placed on the loss term for SOC. Moreover, the model predicts capacity with high precision, which in turn improves SOH prediction using the predicted capacity. Interestingly, the model's direct SOH output is somewhat less accurate compared to $SOH_{\text{cali}}$, possibly due to the more concentrated distribution of SOH values in the dataset compared to capacity. Nevertheless, both methods for predicting SOH yield an acceptable error margin, generally within 3%.

27	B. Saha & Goebel, K.   (ed NASA Ames Research Center NASA Prognostics Data Repository, Moffett Field, CA) (2007).
28	Chen, R. T., Rubanova, Y., Bettencourt, J. & Duvenaud, D. K. Neural ordinary differential equations. *Advances in neural information processing systems* **31** (2018).
29	Pepe, S., Liu, J., Quattrocchi, E. & Ciucci, F. Neural ordinary differential equations and recurrent neural networks for predicting the state of health of batteries. *Journal of Energy Storage* **50**, 104209 (2022). https://doi.org/https://doi.org/10.1016/j.est.2022.104209
30	Xie, T. & Grossman, J. C. Crystal Graph Convolutional Neural Networks for an Accurate and Interpretable Prediction of Material Properties. *Physical Review Letters* **120**, 145301 (2018). https://doi.org/10.1103/PhysRevLett.120.145301
31	Wang, J. *et al.* A comprehensive transformer-based approach for high-accuracy gas adsorption predictions in metal-organic frameworks. *Nature Communications* **15**, 1904 (2024). https://doi.org/10.1038/s41467-024-46276-x
32	Tian, J., Xiong, R., Shen, W., Lu, J. & Sun, F. Flexible battery state of health and state of charge estimation using partial charging data and deep learning. *Energy Storage Materials* **51**, 372-381 (2022). https://doi.org/https://doi.org/10.1016/j.ensm.2022.06.053